\documentclass[11pt]{article}

\usepackage[preprint]{acl}

\usepackage{times}
\usepackage{latexsym}
\usepackage{amsmath}
\usepackage{amssymb}
\usepackage{float}
\usepackage{multirow}

\usepackage[T1]{fontenc}

\usepackage[utf8]{inputenc}

\usepackage{microtype}

\usepackage{inconsolata}

\usepackage{graphicx}
\usepackage{algorithm}
\usepackage{algorithmic}
\usepackage{booktabs}
\usepackage{subcaption}

\title{SPAR: Experience-Driven Strategy Evolution in Multi-Session B2B Sales}

\author{
  Hangming Zhang\textsuperscript{1}\thanks{Equal contribution.}\thanks{Work done during internship at Alibaba Group.} \quad
  Yao Zhao\textsuperscript{2}\footnotemark[1] \quad
  Haotian Kuang\textsuperscript{2} \\
  \bfseries C. L. Philip Chen\textsuperscript{1} \quad
  Tong Zhang\textsuperscript{1} \quad
  Tengyue Ma\textsuperscript{2}\thanks{Corresponding author.} \\
  \normalsize \textsuperscript{1}South China University of Technology \\
  \textsuperscript{2}Alibaba Group \\
  \normalsize hangmingzhang37@gmail.com,
  philip.chen@ieee.org,
  tony@scut.edu.cn, \\
  \normalsize \{zhaoyao.zhao, kuanghaotian.kht, xianyue.mty\}@taobao.com
}

\begin{document}
\maketitle
\begin{abstract}
B2B outbound sales involves high transaction values and relationship-based trust, so closing a deal typically requires multiple rounds of outreach building on prior context. Existing LLM-based sales agents operate within single sessions and cannot carry forward context or refine strategies from experience. Methods that do learn from experience store lessons as unstructured text and treat all outcomes as equally informative, with no mechanism to distinguish a flawed strategy from an external factor. We propose SPAR, a self-evolving framework that models selling strategies as testable hypotheses and refines them without parameter updates. A double-loop architecture lets an inner loop adapt to individual merchants through episodic memory while an outer loop evolves a shared strategy playbook through causal attribution and hypothesis-driven evolution. In a live A/B test on the 1688 platform with over 200,000 merchants, SPAR achieves +15.8\% conversion lift and +35.8\% high-intent merchant growth. Human evaluation by domain experts rates SPAR at 4.62 out of 5 on overall sales quality, the highest among all methods.
\end{abstract}

\section{Introduction}
\label{sec:intro}

In B2B outbound sales, high transaction values, multi-stakeholder decisions, and relationship-based trust mean that closing a deal typically requires multiple rounds of persistent, targeted outreach. Experienced salespeople handle this by carrying forward context, judging which approaches work, and refining their playbook across merchants. Large language models have driven interest in automating this process \citep{chiu2022salesbot,chang2024injecting,kim-etal-2025-towards}, yet existing systems remain limited to single-session interactions \citep{he2018decoupling,shea2024ace}, leaving three closely linked challenges unaddressed.

\begin{figure}[t]
\centering
\includegraphics[width=0.9\columnwidth]{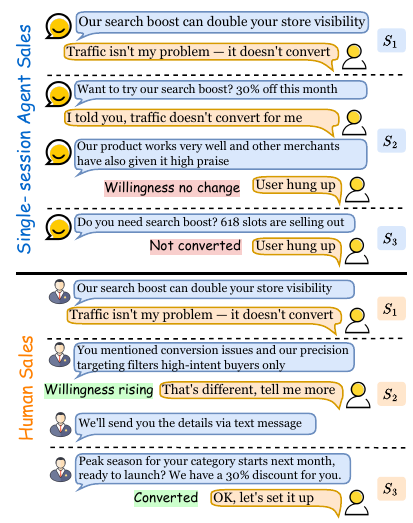}
\caption{Single-session agents repeat generic pitches across isolated calls without addressing the merchant's core objection, leading to repeated rejections. Multi-session human salespeople follow up on specific concerns, build on prior context, and achieve conversion. $S_1, S_2, \ldots$ denote session 1, session 2, and so on.}
\label{fig:intro}
\vspace{-1.5em}
\end{figure}

Current systems treat each session as an isolated event, a \textbf{single-session limitation} that caps conversion at first-contact performance \citep{chang2024injecting}, as Figure~\ref{fig:intro} illustrates. Even with multi-session capability, \textbf{attribution ambiguity} persists, as it remains unclear whether a failed follow-up reflects a flawed strategy or an external factor. This ambiguity entrenches \textbf{manual dependence} on hand-crafted scripts, since automating strategy updates without reliable attribution would amplify rather than correct errors \citep{deng2024plug}. These three barriers operate at successive levels, and each must be resolved for the next to become effective.

While recent LLM-based agents have explored experience-driven self-improvement \citep{shinn2023reflexion,zhao2024expel,park2023generative}, they treat all outcomes as equally informative and store experience as unstructured text, offering no mechanism to filter external noise or to systematically validate and retire strategies.

We propose SPAR (\textbf{S}trategy \textbf{P}laybook via \textbf{A}ttributed \textbf{R}efinement), a self-evolving framework that models selling strategies as \emph{testable hypotheses} and iteratively refines them through sequential merchant interactions without parameter updates. Learning is organized as a double-loop architecture, where an inner loop adapts to individual merchants through multi-session memory, while an outer loop systematically evolves the shared strategy playbook across merchants through causal attribution and hypothesis-driven evolution.

Our contributions are as follows.
\begin{itemize}
    \item We formulate multi-session B2B sales as a double-loop learning problem and propose SPAR, a self-evolving framework that separates within-merchant adaptation from cross-merchant strategy refinement, requiring no parameter updates.
    \item We model selling strategies as testable hypotheses with an evidence-based lifecycle of proposal, testing, validation, and retirement, and introduce causal attribution to isolate genuinely strategy-driven outcomes for reliable playbook updates.
    \item We evaluate SPAR in a live A/B test on the 1688 platform with over 200,000 merchants, achieving +15.8\% conversion lift and +35.8\% high-intent merchant growth. Simulation analysis also shows SPAR outperforms the strongest baseline by 9.6\% on objection recognition and 5.5\% on evidence support.
\end{itemize}

\section{Related Work}
\label{sec:related}

\subsection{Sales Dialogue Agents}

Conversational agents for sales and negotiation \citep{zhan2024let} have progressed from scripted pipelines \citep{chiu2022salesbot} to LLM-powered approaches that inject strategies through fine-tuning \citep{chang2024injecting} and learn adaptive policies \citep{deng2024plug}. Complementary efforts evaluate negotiation capabilities \citep{bianchi2024well} and profile users for personalization \citep{kim-etal-2025-towards,gou2025advancing}. Despite this progress, all existing systems operate within a \textbf{single interaction session}, with no mechanism to revisit merchants or refine strategies from ongoing experience.

Multi-session dialogue has been studied in open-domain settings \citep{xu2022beyond,maharana2024evaluating,li2025hello}. Memory systems and benchmarks built for multi-session agents follow the same framing, where MemMachine \citep{wang2026memmachinegroundtruthpreservingmemorypersonalized} preserves one user's conversational record at the episode level for cheap personalized retrieval and MemoryArena \citep{he2026memoryarena} tests whether an agent uses what earlier sessions gave it rather than recalling it on demand. This line of work is organized around the record of a single user and improves what an agent can recall, but these systems maintain conversational consistency rather than adapt strategically toward conversion.

\subsection{Outcome Attribution}

Determining which action caused an observed outcome is fundamental to learning from sequential interactions. Counterfactual credit assignment in reinforcement learning \citep{mesnard2021counterfactual} distinguishes the agent's contribution from external factors. Despite this importance, LLM agent learning frameworks treat all results as equally informative. SPAR bridges this gap by separating strategy-driven outcomes from externally-determined ones before updating the playbook.

\subsection{Experience-Driven Self-Improvement}

Building on foundational agent frameworks \citep{yao2022react}, recent work enables LLM agents to learn from experience without parameter updates through self-reflection \citep{shinn2023reflexion,madaan2023self,park2023generative}, cross-episode transfer \citep{zhao2024expel,fu2024autoguide,wang2024voyager}, self-generated examples \citep{sarukkai2026self}, and evolving context \citep{zhang2026agentic}. However, these methods store experience as unstructured text or flat rules, with no lifecycle to promote validated strategies and retire failing ones.

\section{Method}
\label{sec:method}

\subsection{Task Formulation}
\label{sec:task}

The agent contacts $N$ merchants sequentially, each with behavioral profile $p_i$. For merchant $i$, the agent conducts up to $K$ dialogue sessions. Session $k$ produces outcome $o_{i,k} \in \{1, 0, -1\}$, representing success, ongoing, and failure respectively. The agent maintains two evolving knowledge stores. An \textbf{episodic memory} $M_i$ for each merchant records structured session-level notes across follow-ups. A shared \textbf{strategy playbook} $\mathcal{P} = \{h_1, \dots, h_m\}$ accumulates selling strategies refined through experience across all merchants. The objective is to maximize the cumulative conversion rate
\begin{equation}
R = \frac{1}{N} \sum_{i=1}^{N}
  \max_{k \leq K}\, \mathbf{1}[o_{i,k} = 1]
\end{equation}
while simultaneously improving $\mathcal{P}$ such that later merchants benefit from validated lessons learned from earlier ones.

\subsection{Framework Overview}
\label{sec:overview}

\begin{figure*}[!t]
\centering
\begin{subfigure}[t]{\textwidth}
  \centering
  \includegraphics[width=\textwidth]{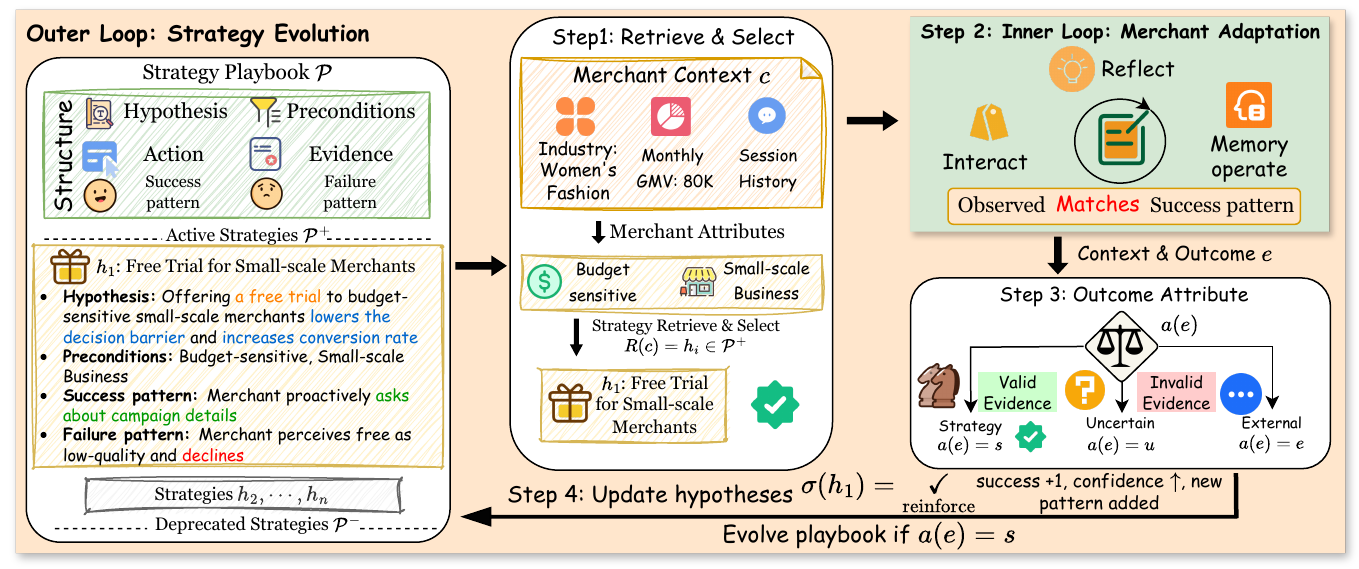}
  \caption{The double-loop architecture. The outer loop maintains and evolves a strategy playbook $\mathcal{P}$ of testable hypotheses through four steps after each interaction. The inner loop adapts to individual merchants across sessions using episodic memory $M_i$.}
  \label{fig:framework-overview}
\end{subfigure}

\vspace{0.5em}

\begin{subfigure}[t]{\textwidth}
  \centering
  \includegraphics[width=\textwidth]{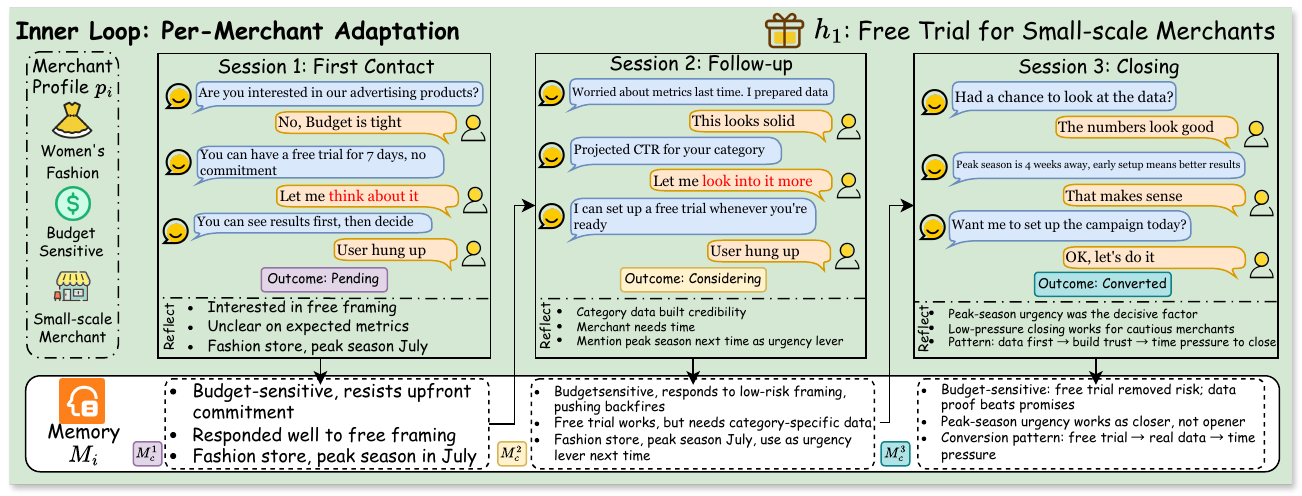}
  \caption{Detailed view of the inner loop for a single merchant across three sessions. Episodic memory accumulates observations from each session and guides follow-up strategies, enabling the agent to build rapport and adapt its approach over time.}
  \label{fig:framework-inner}
\end{subfigure}
\caption{Overview of the SPAR framework. All strategies and merchant data shown are anonymized. The illustration is a simplified example for readability. Real B2B sales scenarios involve more complex merchant profiles and longer interaction sequences.}
\label{fig:framework}
\vspace{-1em}
\end{figure*}

SPAR organizes learning as a \textbf{double-loop architecture}, illustrated in Figure~\ref{fig:framework}. The \textbf{inner loop} adapts to individual merchants across multiple sessions using episodic memory $M_i$. The \textbf{outer loop} evolves $\mathcal{P}$ across merchants by treating each interaction as a natural experiment that tests and refines strategy hypotheses. Algorithm~\ref{alg:spar} summarizes the complete pipeline, and the following subsections detail each component.

\subsection{Strategy as Testable Hypothesis}
\label{sec:hypothesis}

We model each strategy as a \textbf{testable hypothesis} with accumulated evidence that drives lifecycle transitions. The strategy playbook $\mathcal{P}$ is a collection of such hypotheses, each evolved through reinforcement, refinement, addition, merging, and deprecation (\S\ref{sec:outer}). Each hypothesis $h \in \mathcal{P}$ carries an evidence set $E(h)$, where each entry $e \in E(h)$ records the merchant context and interaction outcome. A causal attribution function $a(e) \in \{s, x, u\}$ then labels the outcome as \textbf{s}trategy-driven, e\textbf{x}ternally-determined, or \textbf{u}ncertain. Let $E_s(h) = \{e \in E(h) \mid a(e) = s\}$ denote the strategy-attributed evidence and $E_s^{\mathrm{succ}}(h)$ the subset with successful outcomes. The \textbf{confidence} of $h$ is computed as the posterior mean of a Beta-Bernoulli model with uniform prior
\begin{equation}
c(h) = \frac{1 + |E_s^{\mathrm{succ}}(h)|}{2 + |E_s(h)|}
\label{eq:conf}
\end{equation}
which starts at $c(h) = 0.5$, avoids overconfidence from small samples, and converges to the empirical success rate as evidence accumulates, isolating the causal effect of the strategy by excluding externally-attributed evidence. Each hypothesis has status $\sigma(h) \in \{0, +1, -1\}$ for tentative, active, and deprecated, transitioning by
\begin{equation}
\sigma(h) \leftarrow \begin{cases}
  +1 & |E_s(h)| \geq \tau_p \;\land\; c(h) \geq \theta_p \\
  -1 & |E_s(h)| \geq \tau_d \;\land\; c(h) < \theta_d \\
  \sigma(h) & \text{otherwise}
\end{cases}
\label{eq:lifecycle}
\end{equation}
where $(\tau_p, \theta_p)$ and $(\tau_d, \theta_d)$ are promotion and deprecation thresholds. $\mathcal{P}^{+} = \{h \in \mathcal{P} \mid \sigma(h) \geq 0\}$ denotes the non-deprecated hypotheses. To balance exploitation and exploration, each hypothesis carries an \textbf{exploration priority} combining Bernoulli outcome variance with a sample-size exploration bonus
\begin{equation}
I(h) = c(h) \cdot \bigl(1 - c(h)\bigr)
       + \frac{1}{1 + n}
\label{eq:igp}
\end{equation}
where $n = |E_s(h)|$.

\subsection{Within-Merchant Adaptation}
\label{sec:inner}

The inner loop maintains an \textbf{episodic memory} $M_i$ for each merchant. After each session, the agent generates a structured \textbf{reflection} summarizing what happened, why the outcome occurred, and what to try next, then stores it in $M_i$ alongside the outcome and strategy used. This reflection provides actionable context for follow-up sessions and supplies the qualitative signal needed for causal attribution in the outer loop.

At the start of each session, the agent concatenates the merchant profile $p_i$, interaction history $M_i$, and the non-deprecated hypothesis set $\mathcal{P}^{+}$ as selection context. An LLM-based selector implements the scoring function $f$ in Algorithm~\ref{alg:spar}, choosing $h^{*} \in \mathcal{P}^{+}$ by evaluating precondition alignment, consistency with prior outcomes, and exploration priority $I(h)$. In follow-up sessions, recorded failure modes steer the selector away from repeating unsuccessful strategies.

\subsection{Cross-Merchant Strategy Evolution}
\label{sec:outer}

After each merchant interaction, the outer loop transforms raw experience into systematic playbook improvements through causal attribution, hypothesis evolution, and quality curation.

\paragraph{Causal Attribution.} The LLM examines the dialogue, merchant profile, outcome, and strategy used, then assigns an attribution label $a(e) \in \{s, x, u\}$. This counterfactual credit assignment \citep{mesnard2021counterfactual} gates playbook updates. Only outcomes with $a(e) = s$ enter $E_s(h^{*})$ and trigger evolution. External outcomes are recorded but do not affect $c(h)$.

\paragraph{Hypothesis Evolution.} For outcomes causally attributed to the strategy, the agent applies structured delta updates to the relevant hypothesis
\begin{equation}
\mathcal{P} \leftarrow \Phi\!\left(\mathcal{P},\; h^{*},\; e\right)
\end{equation}
where $e$ is the newly constructed evidence. The LLM implements $\Phi$ by selecting from seven evolution actions, designed around a \textbf{minimal revision principle} that always prefers the smallest targeted change sufficient to incorporate new evidence. \emph{Reinforce}, \emph{Update}, \emph{Condition}, and \emph{None} preserve existing hypotheses with increasing degrees of modification. \emph{Add}, \emph{Merge}, and \emph{Deprecate} restructure the playbook more aggressively when incremental updates are insufficient. Together these complementary actions cover evidence accumulation, hypothesis refinement, and structural reorganization.

\paragraph{Exploration and Curation.} The agent performs \textbf{fork discovery} after each interaction, identifying unexplored strategy directions from the dialogue as exploration candidates. The playbook is deduplicated at regular intervals, and the deprecation rule in Eq.~\ref{eq:lifecycle} is applied after every interaction.

\section{Experiments}
\label{sec:experiments}

\subsection{Experimental Setup}
\label{sec:setup}

\paragraph{Dataset.} We construct a multi-session B2B sales evaluation dataset from 1,642 real conversations between an experienced human sales team and merchants on 1688, a major B2B online platform. Each conversation is manually annotated with detailed merchant profiles, behavioral archetypes, session-level outcomes, and strategy labels. 

\paragraph{Merchant Simulator.} We use GPT-4o as the simulated merchant, conditioned on a detailed profile with target prices, behavioral traits, and session-specific goals. We validate fidelity by replaying 300 real conversations and conducting a human evaluation with the sales team. The simulator is conditioned on the full conversation context from real human merchant interactions, yielding 98.7\% outcome agreement as shown in Table~\ref{tab:fidelity}. The 87\% paired comparison accuracy further confirms that the simulator responds differently to strong and weak strategies. Expert approval reaches 89\%, validating its use as a reliable evaluation proxy.

\begin{table}[t]
\centering
\small
\begin{tabular}{@{}ll@{}}
\toprule
Metric & Score \\
\midrule
Outcome agreement & 98.7\% \\
Key-behavior recall & 91\% \\
Turn-count deviation & 0.9 $\pm$ 1.3 \\
Paired comparison accuracy & 87\% \\
Expert approval & 89\% \\
\bottomrule
\end{tabular}
\caption{Merchant simulator fidelity validated against 300 real conversations and human evaluation from the sales team.}
\label{tab:fidelity}
\vspace{-1.5em}
\end{table}

\paragraph{Baselines.} We compare SPAR against three baselines. \textbf{Direct Prompting} uses the same backbone LLM with only the item description and no memory or evolving playbook. \textbf{Reflexion} \citep{shinn2023reflexion} generates a natural-language lesson after each conversation and stores it per merchant as self-reflective feedback for future interactions. \textbf{ExpeL} \citep{zhao2024expel} extracts reusable insights from past interactions and retrieves them based on task similarity to guide decision-making.

\paragraph{Metrics.} We evaluate agents along three dimensions of sales dialogue quality.
\begin{itemize}
    \item \textbf{Objection handling} measures whether the agent recognizes the merchant's core objection, responds to it directly, and supports its response with concrete evidence.
    \item \textbf{Dialogue quality} is assessed through \emph{monologue rate}, the fraction of sessions dominated by lengthy seller utterances, and \emph{repetition rate}, the fraction of follow-up sessions that duplicate earlier questions or arguments.
    \item \textbf{Memory recall} measures whether the agent proactively references specific facts from prior sessions before the merchant raises them, evaluated only after the first session.
\end{itemize}
\vspace{-0.5em}
For LLM-based evaluation, all metrics except repetition rate are scored by an LLM evaluator \citep{zheng2023judging} at the session level across all real conversations. Repetition rate is computed via text similarity between consecutive sessions.

\paragraph{Hypothesis Design.} The hypothesis schema is derived from LLM-assisted analysis of 51,603 session notes from a top-performing B2B sales team on 1688 and validated against 3,637 sales cases. Each hypothesis contains six fields: \emph{description}, \emph{preconditions}, \emph{action steps}, \emph{rationale}, \emph{success and failure patterns}, and an \emph{evidence} set with causal attribution labels. Confidence scoring and lifecycle thresholds are similarly grounded in observed patterns of how experienced salespeople adopt and retire strategies. The session notes used for schema design have no overlap with the 1,642 evaluation conversations.

\paragraph{Implementation.} All agents use GPT-4o as the backbone LLM. The playbook is initialized with a small set of domain-specific tentative seed strategies and capped at 50 entries.

\begin{table*}[!t]
\centering
\footnotesize
\setlength{\tabcolsep}{4pt}
\renewcommand{\arraystretch}{0.9}
\begin{tabular}{c l ccc cc c c}
\toprule
& & \multicolumn{3}{c}{Objection Handling} & \multicolumn{2}{c}{Dialogue Quality} & Memory & Sales Quality \\
\cmidrule(lr){3-5} \cmidrule(lr){6-7} \cmidrule(lr){8-8} \cmidrule(l){9-9}
& Method & Recognize$\uparrow$ & Respond$\uparrow$ & Evidence$\uparrow$ & Monologue$\downarrow$ & Repetition$\downarrow$ & Recall$\uparrow$ & Score$\uparrow$ \\
\midrule
\multirow{7}{*}[-1ex]{\rotatebox[origin=c]{90}{\scriptsize GPT Eval}}
& Direct Prompting & 0.741 & 0.540 & 0.428 & \underline{\textbf{0.186}} & \underline{0.005} & -- & -- \\
& ExpeL & 0.705 & 0.623 & 0.607 & 0.262 & 0.016 & -- & -- \\
& Reflexion & 0.759 & 0.638 & \underline{0.672} & \underline{0.221} & 0.017 & 0.053 & -- \\
\cmidrule(l){2-9}
& SPAR w/o Inner Loop & 0.684 & 0.579 & 0.596 & 0.281 & 0.018 & -- & -- \\
& SPAR w/o Outer Loop & \underline{0.800} & \underline{0.672} & \underline{0.672} & 0.272 & 0.009 & \underline{0.065} & -- \\
& SPAR w/o Evolution & 0.695 & 0.661 & 0.627 & 0.288 & 0.007 & -- & -- \\
\cmidrule(l){2-9}
& SPAR & \underline{\textbf{0.832}} & \underline{\textbf{0.729}} & \underline{\textbf{0.709}} & \underline{0.221} & \underline{\textbf{0.002}} & \underline{\textbf{0.071}} & -- \\
\midrule
\multirow{7}{*}[-1ex]{\rotatebox[origin=c]{90}{\scriptsize Claude Eval}}
& Direct Prompting & 0.714 & 0.492 & 0.415 & \underline{\textbf{0.178}} & \underline{0.006} & -- & -- \\
& ExpeL & 0.642 & 0.578 & 0.548 & 0.275 & 0.020 & -- & -- \\
& Reflexion & 0.728 & 0.584 & 0.621 & 0.211 & 0.024 & 0.042 & -- \\
\cmidrule(l){2-9}
& SPAR w/o Inner Loop & 0.635 & 0.542 & 0.536 & 0.289 & 0.018 & -- & -- \\
& SPAR w/o Outer Loop & \underline{0.761} & 0.618 & \underline{0.628} & 0.252 & 0.013 & \underline{0.059} & -- \\
& SPAR w/o Evolution & 0.651 & \underline{0.622} & 0.574 & 0.284 & 0.009 & -- & -- \\
\cmidrule(l){2-9}
& SPAR & \underline{\textbf{0.798}} & \underline{\textbf{0.681}} & \underline{\textbf{0.662}} & \underline{0.205} & \underline{\textbf{0.003}} & \underline{\textbf{0.064}} & -- \\
\midrule
\multirow{4}{*}{\rotatebox[origin=c]{90}{\tiny Human Eval}}
& Direct Prompting & 0.762 & 0.884 & 0.741 & \underline{0.306} & 0.346 & -- & 2.83 \\
& ExpeL & \underline{0.871} & 0.857 & 0.879 & 0.384 & 0.337 & -- & 3.64 \\
& Reflexion & 0.846 & \underline{0.903} & \underline{0.896} & 0.318 & \underline{0.162} & \underline{0.762} & \underline{3.97} \\
& SPAR & \underline{\textbf{0.921}} & \underline{\textbf{0.938}} & \underline{\textbf{0.926}} & \underline{\textbf{0.258}} & \underline{\textbf{0.132}} & \underline{\textbf{0.813}} & \underline{\textbf{4.62}} \\
\bottomrule
\end{tabular}
\caption{Results on the B2B sales evaluation dataset. \underline{\textbf{Bold with underline}} marks the best and \underline{underline} marks the second-best in each column. Recognize, Respond, and Evidence measure objection handling, Monologue and Repetition measure dialogue quality, and Recall measures proactive memory callback in follow-up sessions. GPT Eval and Claude Eval score the same dialogues with GPT-4o and with Claude Sonnet 4.5. Sales Quality is rated by three domain experts on a 1--5 scale under blind conditions (rubric in Appendix~\ref{sec:appendix-rubric}).}
\label{tab:main}
\vspace{-1.5em}
\end{table*}

\subsection{Results and Ablation}
\label{sec:main}

\paragraph{LLM-based Evaluation.} Table~\ref{tab:main} presents results across baselines and ablation variants under two judge models, where SPAR \textbf{leads on most metrics} under both. Under GPT Eval, the advantage is most pronounced on \textbf{evidence support}, where SPAR achieves \textbf{65.6\% higher} than Direct Prompting and \textbf{5.5\% higher} than the strongest baseline, showing that the double-loop architecture yields more substantive selling arguments. SPAR also achieves the lowest repetition rate at 0.002 and the highest memory recall at 0.071. ExpeL and Reflexion improve through experience accumulation, yet both lack the structured hypothesis lifecycle that allows SPAR to validate effective strategies and retire failing ones. The recognition gain over the strongest baseline is \textbf{9.6\%}. Under Claude Eval, SPAR leads the strongest baseline by \textbf{16.6\%} on response, cuts the repetition rate to \textbf{0.003} from 0.006, and raises memory recall to \textbf{0.064} from 0.042.

\paragraph{Human Evaluation.} Three domain experts annotate dialogues with 300+ real merchants under blind conditions (Table~\ref{tab:main}). SPAR achieves the \textbf{highest scores across all seven metrics}, with a Sales Quality of \textbf{4.62 out of 5} and evidence support at \textbf{0.926}. The advantage is most visible on \textbf{repetition} at \textbf{0.132}, an \textbf{18.5\%} reduction over the next-best method, confirming that playbook evolution produces non-repetitive follow-up interactions. Direct Prompting scores competitively on response at \textbf{0.884}, reflecting the strong zero-shot capability of the backbone LLM, but falls behind on recognition at \textbf{0.762} and evidence at \textbf{0.741}, where structured learning provides the largest gains. The performance gaps are narrower than in LLM-based evaluation under both judge models, yet the \textbf{ranking across methods remains consistent}, confirming that the automated metrics reliably capture quality differences perceived by experienced salespeople.

\paragraph{Ablation Study.} The ablation variants isolate each component's contribution. Under GPT Eval, removing the inner loop causes the \textbf{largest degradation}, with recognition dropping \textbf{17.8\%} and memory recall falling to zero, showing that \textbf{per-merchant memory is essential} for cross-session continuity. Removing the outer loop retains strong recognition at only 3.8\% below SPAR but lags \textbf{7.8\%} on response and \textbf{5.2\%} on evidence, confirming that \textbf{the evolving playbook drives deeper objection handling}. Removing evolution freezes the playbook after seeding, falling \textbf{9.3\%} behind on response and \textbf{11.6\%} on evidence, as the agent cannot refine strategies from new experience. Under Claude Eval, recognition falls \textbf{20.4\%} without the inner loop and evidence falls \textbf{13.3\%} without evolution. These results confirm that \textbf{all three components are necessary and mutually reinforcing}.

\begin{table}[t]
\centering
\small
\begin{tabular}{@{}lccc@{}}
\toprule
Method & Conv.$\uparrow$ & Sessions$\downarrow$ & M1 Rate$\uparrow$ \\
\midrule
Direct Prompting & -- & -- & -- \\
ExpeL & +1.1\% & +0.3\% & +9.2\% \\
Reflexion & +9.6\% & $-$4.6\% & +27.9\% \\
\midrule
SPAR w/o Inner Loop & +10.1\% & $-$6.3\% & \underline{+34.5\%} \\
SPAR w/o Outer Loop & \underline{+12.6\%} & \underline{\textbf{$-$11.2\%}} & +22.3\% \\
SPAR w/o Evolution & +9.6\% & $-$3.0\% & +18.7\% \\
\midrule
SPAR & \underline{\textbf{+15.8\%}} & \underline{$-$9.2\%} & \underline{\textbf{+35.8\%}} \\
\bottomrule
\end{tabular}
\caption{Online A/B test results on the 1688 platform. All values are relative changes over the Direct Prompting baseline. Conv. is conversion rate, Sessions is the number of outreach sessions needed, and M1 Rate is first-milestone achievement rate. \underline{\textbf{Bold with underline}} marks the best and \underline{underline} marks the second-best.}
\label{tab:online}
\vspace{-1.5em}
\end{table}

\subsection{Online A/B Test}
\label{sec:online}

To validate that simulation improvements transfer to the real world, we deploy all methods on the 1688 platform in a live A/B test with over 30,000 merchants per group. Conversion is defined as completing a target transaction within 7 days of the final outreach session. The M1 high-intent rate measures how often merchants reach high-intent status during the sales process. Table~\ref{tab:online} reports the relative change over the Direct Prompting baseline on three business metrics. Statistical significance is assessed via a two-proportion z-test with 95\% CIs from 1{,}000-iteration bootstrap, and all SPAR lifts are significant at $p < 0.001$.

SPAR achieves the \textbf{highest conversion lift at +15.8\%} and the \textbf{highest M1 achievement lift at +35.8\%}, while reducing outreach sessions by \textbf{9.2\%}. The conversion advantage over the next-best variant is \textbf{25.4\%}, confirming that the full double-loop architecture delivers the strongest end-to-end selling capability in production. At scale, this yields operational savings while easing merchant fatigue from repeated contact.

The ablation pattern mirrors the simulation results. SPAR w/o Outer Loop achieves the second-highest conversion at +12.6\% and the largest session reduction at $-$11.2\%, while SPAR w/o Inner Loop yields +10.1\% conversion with the highest M1 lift among ablation variants at +34.5\%. SPAR w/o Evolution drops M1 to +18.7\%, confirming that \textbf{strategy evolution nearly doubles the M1 gain}. ExpeL shows minimal improvement at +1.1\% conversion, suggesting that unstructured insight accumulation provides limited benefit in multi-session B2B sales. We provide a qualitative analysis of playbook evolution in Appendix~\ref{sec:appendix-evolution}.

\section{Conclusion}
\label{sec:conclusion}

We presented SPAR, a self-evolving framework that models selling strategies as testable hypotheses and refines them through a double-loop architecture with causal attribution. In a live A/B test on the 1688 platform with over 200,000 merchants, SPAR achieves +15.8\% conversion lift and +35.8\% high-intent merchant growth. Simulation analysis shows SPAR outperforms the strongest baseline by 9.6\% on objection recognition and 5.5\% on evidence support. These results are validated in real-world B2B commercial outreach where merchants make purchasing decisions.

\section*{Limitations}

SPAR requires multiple interactions before the playbook meaningfully evolves, so early merchants do not benefit from accumulated experience. This cold-start cost is inherent to any online learning approach, and investigating warm-start techniques such as transferring playbooks across domains is a promising direction for future work.

The strategy evolution process relies on LLM reasoning, which inherently introduces a degree of randomness and unpredictability in playbook updates. While the confidence mechanism and lifecycle thresholds provide gradual self-correction over time, individual evolution decisions may be suboptimal. Incorporating an auxiliary verification agent to validate proposed strategy changes before they enter the playbook is a promising direction for improving reliability.

Merchants are processed sequentially in our current experiments. In practice, sales teams handle multiple merchants concurrently, where parallel interactions may produce fundamentally different learning dynamics. Evaluating SPAR in concurrent settings remains future work.

\section*{Acknowledgments}

This work is supported by the CCF-1688 Yuanbao Cooperation Fund (No. CCF-Alibaba2025007) and the Key-Area Research and Development Program of Guangdong Province (No. 2023B0303030001).

\bibliography{custom}

\appendix

\section{Playbook Evolution Analysis}
\label{sec:appendix-evolution}

We examine how the strategy playbook evolves during the online A/B test to illustrate the hypothesis lifecycle. The playbook is initialized with six tentative seed strategies. As larger-scale manual analysis is infeasible, we qualitatively examine its evolution on a sample of 200+ real merchants drawn from the A/B test.

\paragraph{Lifecycle Progression.} Two of the six seed strategies are promoted to \emph{active} status through accumulated evidence. Strategy H002, which instructs the agent to reference prior conversations with returning merchants, accumulates 108 strategy-attributed successes and is promoted after reaching the evidence threshold. Strategy H001, which recommends offering trial programs when merchants hesitate, follows a similar trajectory with 89 attributed successes. Two other seed strategies are deprecated after repeated failures, illustrating that the lifecycle correctly retires ineffective approaches. The evolution process also discovers and validates five new strategies beyond the initial seeds, demonstrating that the playbook grows organically from interaction experience.

\paragraph{Fork Discovery.} The evolution process identifies 9 unexplored strategy directions from dialogue analysis. These directions capture recurring patterns that existing strategies do not address, such as offering hands-on onboarding support and providing industry-specific case studies to build credibility. Each direction is logged with the number of times it appears across interactions, creating a prioritized queue for future hypothesis generation.

\paragraph{Attribution Filtering.} The causal attribution mechanism labels a subset of outcomes as externally driven, preventing them from distorting strategy confidence scores. When a merchant walks away due to a recently reduced internal budget rather than a flaw in the agent's approach, the attribution module assigns label $x$ and the corresponding evidence does not affect $c(h)$. This filtering ensures that the playbook reflects genuine strategy effectiveness rather than environmental noise.

\section{Algorithm}
\label{sec:appendix-algo}

Algorithm~\ref{alg:spar} describes the complete SPAR pipeline. The outer loop iterates over merchants sequentially. For each merchant, the inner loop runs up to $K$ sessions. At each session, the agent selects a strategy hypothesis $h^{*}$ from the non-deprecated set $\mathcal{P}^{+}$ using a scoring function that considers precondition alignment, prior interaction history, and exploration priority $I(h)$. After dialogue execution, a reflection is stored in episodic memory $M_i$. The outer loop then constructs evidence from the interaction, assigns a causal attribution label, and triggers playbook evolution $\Phi$ only for strategy-attributed outcomes. The session loop terminates early on deal closure ($o=1$) or merchant decline with no follow-up opportunity ($o=-1$).

\begin{algorithm}[H]
\caption{SPAR Double-Loop Learning}
\label{alg:spar}
\begin{algorithmic}[1]
\REQUIRE Merchants $1, \dots, N$ with profiles $\{p_i\}$, max sessions $K$, playbook $\mathcal{P}$
\ENSURE Evolved playbook $\mathcal{P}$
\FOR{$i = 1$ \TO $N$}
  \STATE Initialize episodic memory $M_i \leftarrow \emptyset$
  \FOR{$k = 1$ \TO $K$}
    \STATE \textit{// Inner Loop - Strategy Selection}
    \STATE $h^{*} \leftarrow \arg\max_{h \in \mathcal{P}^{+}} f(h,\, p_i,\, M_i,\, I(h))$
    \STATE Conduct dialogue using $h^{*}$, observe $o_{i,k}$
    \STATE Generate reflection, store in $M_i$
    \STATE \textit{// Outer Loop - Evolution}
    \STATE Construct evidence $e$ from $p_i$, $h^{*}$, $o_{i,k}$, and dialogue
    \STATE Assign attribution $a(e) \in \{s, x, u\}$
    \STATE Append $e$ to evidence set $E(h^{*})$
    \IF{$a(e) = s$}
      \STATE $\mathcal{P} \leftarrow \Phi(\mathcal{P},\, h^{*},\, e)$
    \ENDIF
    \STATE Update $\sigma(h)$ for all $h \in \mathcal{P}$
    \IF{$o_{i,k} = 1$}
      \STATE \textbf{break} \hfill \textit{// deal closed}
    \ELSIF{$o_{i,k} = -1$}
      \STATE \textbf{break} \hfill \textit{// declined, no follow-up}
    \ENDIF
  \ENDFOR
\ENDFOR
\RETURN $\mathcal{P}$
\end{algorithmic}
\end{algorithm}

\section{Sales Quality Rubric}
\label{sec:appendix-rubric}

Three domain experts rate each dialogue on a 1--5 scale based on the following rubric. Each level describes both overall dialogue quality and selling-strategy execution along four dimensions, namely \emph{strategic fit}, \emph{objection handling}, \emph{rapport building}, and \emph{closing effectiveness}.
\begin{itemize}
    \item \textbf{5 -- Excellent.} The agent demonstrates outstanding professional handling and effectively guides toward conversion. Strategic fit matches the merchant's profile and stage precisely, objections are addressed with concrete evidence, rapport is consistently built across sessions, and closing is clear and well-timed. Suitable as a reference example for sales training.
    \item \textbf{4 -- Good.} The agent handles the conversation professionally. Strategic fit is generally appropriate, objections are addressed with relevant supporting points, rapport is established across sessions, and closing is clear, with only minor opportunities for improvement.
    \item \textbf{3 -- Acceptable.} The agent completes the basic sales flow. Strategic fit is partial, objections are acknowledged but not always supported with concrete evidence, rapport is limited, and closing lacks decisiveness.
    \item \textbf{2 -- Poor.} The agent makes noticeable errors. Strategic fit is weak or mismatched, key objections are missed or inadequately handled, rapport is minimal, and no clear closing is offered, reducing the likelihood of conversion.
    \item \textbf{1 -- Unacceptable.} The agent provides irrelevant or inaccurate responses. Strategic fit is broken, objections are ignored or aggravated, no rapport is established, and the conversation may harm the merchant relationship.
\end{itemize}

\end{document}